\documentclass[runningheads]{llncs}
\usepackage[T1]{fontenc}
\usepackage{graphicx}
\usepackage{amsmath}
 \usepackage{booktabs}
 \usepackage{multirow}
 \usepackage{subfig}
 \usepackage{hyperref}
\begin{document}
\title{ICDAR 2025 Competition on FEw-Shot Text line segmentation of ancient handwritten documents (FEST)}
\titlerunning{ICDAR 2025 FEST Competition}
%

\author{Silvia Zottin$^{*,}$\inst{1}\orcidID{0000-0003-0820-7260} \and
        Axel {De Nardin}$^{*,}$\inst{1}\orcidID{0000-0002-0762-708X} \and
        Giuseppe Branca\inst{1}
        \and
        Claudio Piciarelli\inst{1}\orcidID{0000-0001-5305-1520} \and 
        {Gian Luca} Foresti\inst{1}\orcidID{0000-0002-8425-6892}
        }
\authorrunning{S. Zottin et al.}
%
\def\thefootnote{*}\footnotetext{These authors contributed equally to this work}\def\thefootnote{\arabic{footnote}}

\institute{Department of Mathematics, Computer Science and Physics, University of Udine, Udine, Italy\\
\email{branca.giuseppe@spes.uniud.it}\\
\email{\{silvia.zottin, axel.denardin, claudio.piciarelli, gianluca.foresti\}@uniud.it}}

\maketitle              
\begin{abstract}
Text line segmentation is a critical step in handwritten document image analysis. Segmenting text lines in historical handwritten documents, however, presents unique challenges due to irregular handwriting, faded ink, and complex layouts with overlapping lines and non-linear text flow. Furthermore, the scarcity of large annotated datasets renders fully supervised learning approaches impractical for such materials.
To address these challenges, we introduce the Few-Shot Text Line Segmentation of Ancient Handwritten Documents (FEST) Competition. Participants are tasked with developing systems capable of segmenting text lines in U-DIADS-TL dataset, using only three annotated images per manuscript for training. The competition dataset features a diverse collection of ancient manuscripts exhibiting a wide range of layouts, degradation levels, and non-standard formatting, closely reflecting real-world conditions.
By emphasizing few-shot learning, FEST competition aims to promote the development of robust and adaptable methods that can be employed by humanities scholars with minimal manual annotation effort, thus fostering broader adoption of automated document analysis tools in historical research.

\end{abstract}
\section{Introduction}

Text line segmentation holds fundamental importance among the phases of Document Image Analysis and is the subject of significant and ongoing research. It aims to identify the logical structure of a document. 
This step is crucial because it directly impacts the quality of subsequent processes, such as optical word or character recognition, layout analysis, and information extraction. In the context of ancient and handwritten documents, text line segmentation becomes even more challenging due to various factors like irregular handwriting styles, faded ink, degraded paper quality, and the lack of consistent formatting. These documents often exhibit complex layouts, including overlapping lines, non-linear text flow, and variations in line spacing, making automated segmentation difficult to achieve with high accuracy. 
In traditional segmentation tasks, large annotated datasets are typically required to train machine learning models effectively. However, in historical and handwritten document analysis, obtaining such datasets is often impractical due to the scarcity and uniqueness of the source material.

For these reasons, we propose the Competition on FEw-Shot Text line segmentation of ancient handwritten documents (FEST).

To achieve this goal, we challenge researchers to develop a framework capable of accurately identifying text lines in ancient manuscript pages with pixel-level precision. To make the challenge more interesting, we built a novel dataset consisting of multi-language, multi-column manuscripts with very heterogeneous layouts to mimic a real-world application scenario as closely as possible.
Furthermore, we challenge the participants to perform the task of text line segmentation on the proposed dataset in a few-shot setting, providing only three images, with their corresponding ground truths, as the training set for each one of these manuscripts comprising it.
The ICDAR 2024 Competition on Few-Shot and Many-Shot Layout Segmentation of Ancient Manuscripts (SAM)~\cite{competition} recently addressed the challenge of document layout segmentation under a few-shot learning paradigm, using the U-DIADS-Bib dataset~\cite{Zottin2024}. However, to our knowledge, the FEST Competition is the first to specifically focus on text line segmentation of handwritten historical documents in a few-shot setting and with it, we aim to draw researchers' attention to the importance of making these analysis techniques applicable in the real world. The ability to segment an entire manuscript using only a few manually segmented images for the training process would greatly benefit humanities scholars as it would allow them to employ the system for their studies without the need to employ a significant amount of time producing the ground truths for each manuscript they are interested in. We believe this could help overcome the hurdle of limited adoption within the humanities community.

The description of the ICDAR 2025 FEST competition and the complete regulations can be seen on the dedicated website\footnotemark[1].
\footnotetext[1]{\url{https://ai4ch.uniud.it/FESTcompICDAR25/}}

\begin{figure}[htb]
\centering
    \includegraphics[width=.7\linewidth]{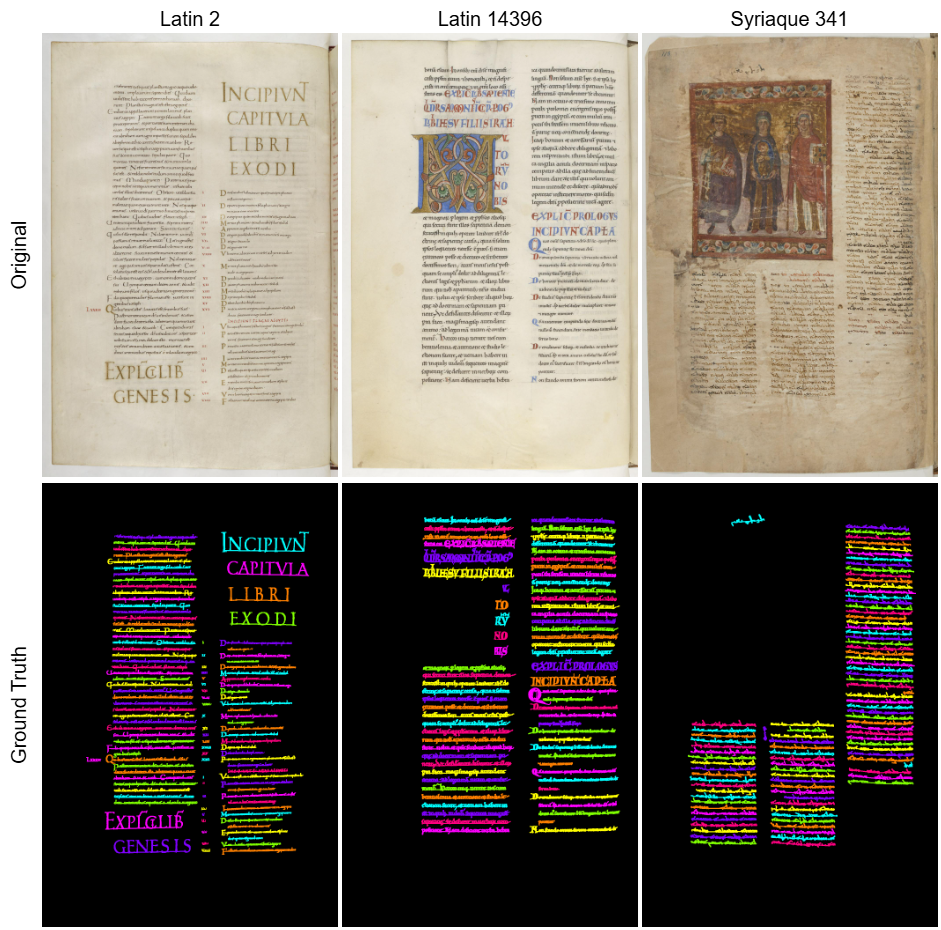} 
    \caption{Original pages and GT samples from the three manuscripts (Latin 2, Latin 14396, and Syriaque 341) included in U-DIADS-TL are presented. In the GT samples, each color denotes a distinct text line segmentation instance.}%
    \label{fig:exampleGT}%
\end{figure}

\section{Dataset}
The dataset used in the competition is the Uniud - Document Image Analysis DataSet - Text Line version (U-DIADS-TL), a proprietary collection developed through a collaboration between computer scientists and humanities scholars at the University of Udine. It consists of 84 color page images selected from three distinct historical manuscripts, Latin 2, Latin 14396, and Syriaque 341, with 28 pages drawn from each. These manuscripts differ significantly in layout complexity, degree of degradation due to preservation conditions and aging, and the alphabet in which they are written, thus presenting a diverse and challenging benchmark for text line segmentation.

Each image is accompanied by a pixel-accurate ground truth (GT) mask in PNG format, matching the original in resolution. The GT annotations comprise two mutually exclusive classes: background and text lines. Compared to existing document segmentation datasets, U-DIADS-TL is distinguished by its fine-grained, noise-free labels, non-ambiguous class regions, multi-column layouts, and heterogeneously oriented text lines, which add to its complexity and research value. Fig.~\ref{fig:exampleGT} illustrates some examples of the defined GT and corresponding original image for each manuscript of the U-DIADS-TL.

For the competition, participants were provided with 13 annotated images per manuscript, three of which constituted the training set. The remaining ten could be used for validation. An additional 15 test images per manuscript were kept private and used for the final evaluation of the submitted models after the challenge concluded.

\begin{figure}[htb]
\centering
    \includegraphics[width=.65\linewidth]{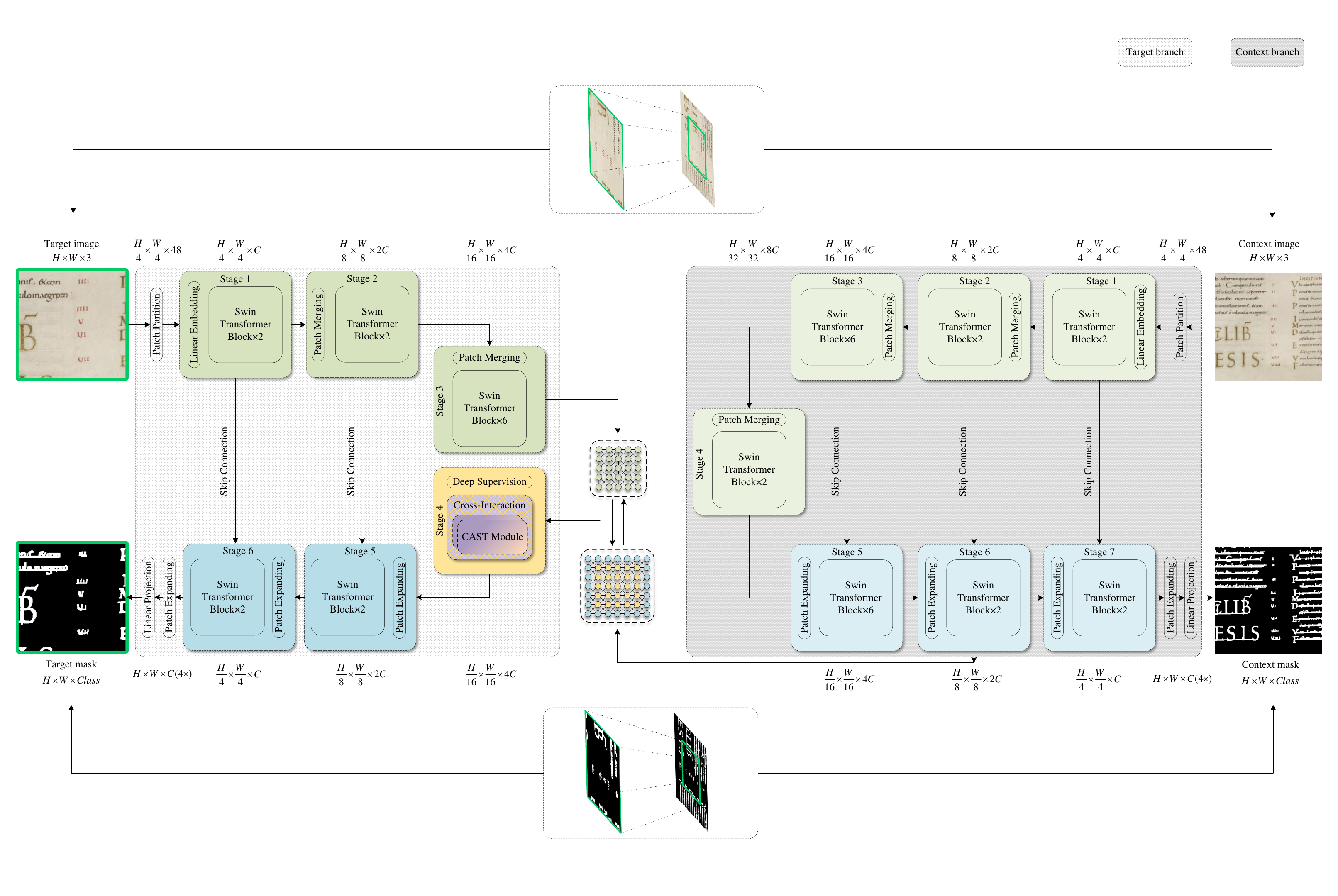} 
    \caption{Architectural details of our HookFormer approach.}%
    \label{imgfei}%
\end{figure}

\section{Systems Description}
In the following section, we briefly introduce the competing systems as described by their respective development teams. A total of 11 teams submitted their results. The participating teams are listed below.

\subsubsection{CV-Group: Fei Wu, and Vincent Christlein, from Computer Vision Group of Pattern Recognition Lab of Friedrich-Alexander-Universität Erlangen-Nürnberg.}

The proposed approach leverages HookFormer~\cite{10440599}, a cutting-edge multiscale Transformer network, to perform semantic segmentation for line extraction tasks. HookFormer adopts a two-branch architecture that integrates both global and local information: the context branch captures coarse, low-resolution features to model structural semantics, while the target branch focuses on fine-grained, high-resolution details for precise pixel-level predictions. A feature hooking mechanism aligns the context and target features, facilitating effective information transfer. To enhance global and local interactions, Cross-Attention Swin-Transformer blocks are employed within a Cross-Interaction module. Figure~\ref{imgfei} illustrates the architecture details of HookFormer model.

Training is stabilized through deep supervision applied to the bottleneck feature maps of the target branch, introducing an additional loss term. The overall loss combines cross-entropy and Dice losses computed on the target branch, context branch, and bottleneck outputs, weighted by hyperparameters to balance contributions.

Each manuscript in the U-DIADS-TL dataset is handled by a dedicated model, trained individually from only three training samples with strong data augmentations (random rotations, flips). Both branches process 224×224 patches, and the network is optimized using stochastic gradient descent with learning rate decay. Swin-Unet backbones, pretrained on ImageNet~\cite{5206848}, serve as the feature extractors for each branch.

Beyond semantic segmentation, the approach incorporates a positional refinement stage. A Mask R-CNN model~\cite{He_2017_ICCV}, pretrained on the READ-BAD dataset~\cite{8395221}, predicts instance-level layouts on the full-page images. These predictions are used to refine the semantic segmentation outputs by correcting fragmented or merged line segments. Horizontal refinement merges broken lines along the horizontal axis, while vertical refinement separates overlapping components vertically, resulting in enhanced line continuity and separation.
The final predictions are assembled by merging patch-based outputs, without additional post-processing steps.

\subsubsection{Codecrackers: Sai Koushik Reddy Yennam, Jaswanth Pederedla, and Tarun Karothi from Indian Institute of Information Technology Sri City, Andhra Pradesh.}

The proposed approach employs a U-Net-based deep learning architecture to address the challenge of text line segmentation in historical documents. U-Net~\cite{unet}, originally developed for biomedical image segmentation, features an encoder-decoder structure with skip connections that effectively combine low-level spatial details with high-level contextual information. The encoder progressively reduces spatial dimensions while increasing feature depth through stacked convolutional and max-pooling layers, capturing hierarchical representations of the document layout. The decoder subsequently reconstructs fine-grained segmentation maps by upsampling feature maps and merging them with corresponding encoder outputs via skip connections, preserving critical spatial information.

All images and masks are resized to 512×512 pixels and normalized to the [0,1] range to standardize inputs. The model is trained with the Adam optimizer (learning rate 0.001) and binary cross-entropy loss, using early stopping and model checkpointing strategies to prevent overfitting.
Training was performed over 47 epochs on a local machine equipped with 16GB RAM.

\begin{figure}[htb]
\centering
\includegraphics[width=\linewidth]{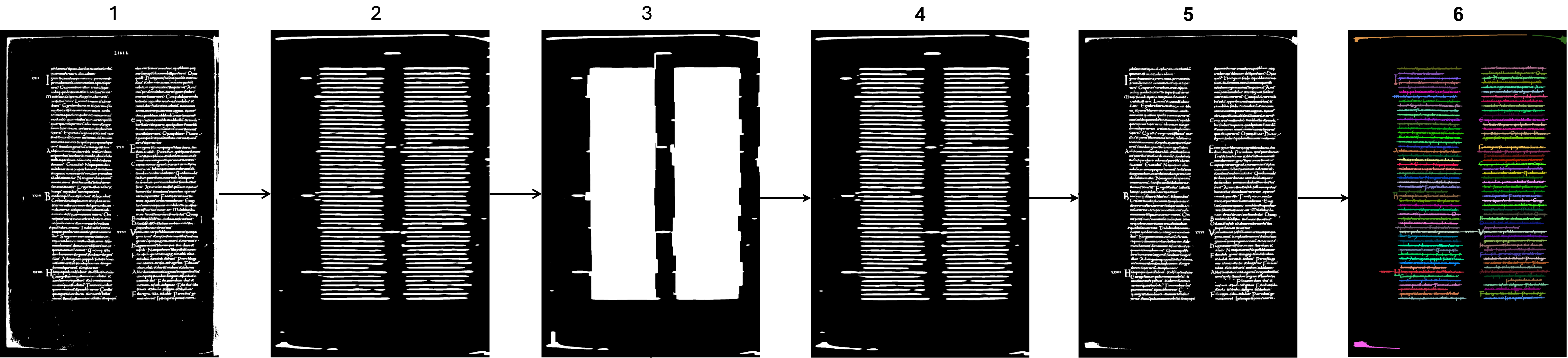} 
    \caption{The TAU-CH pipeline. (1) the grayscale image is binarized and dilated; (2) anisotropic Gaussian filtering at character-height scale detects text lines; (3) a vertical separator mask is generated; (4) it splits merged detections across columns; (5) blobs are vertically elongated to identify text regions; (6) a watershed algorithm, guided by the detected lines, segments individual text lines.}%
    \label{tauch}%
\end{figure}

\subsubsection{TAU-CH: Berat Kurar-Barakat, Sharva Gogawale, Mohammad Suliman, and Nachum Dershowitz, from Tel Aviv University Computational Humanities, Tel Aviv University.}

The proposed method segments text lines in historical document images through a two-stage pipeline (Fig.~\ref{tauch}). In the first stage, the spatial locations of text lines are detected, while in the second stage, the actual text line pixels are segmented using the detected line structures as guidance.

The input grayscale image is initially binarized using Otsu’s thresholding~\cite{otsu}. To enhance line continuity, morphological dilation is applied to the binary image.
An estimate of the average character height range is obtained by analyzing the bounding boxes of connected components that fall within a plausible vertical size interval.
Anisotropic Gaussian filtering is applied with scale parameters derived from the average character height and a fixed elongation factor ($\eta = 3$).An anisotropic Gaussian filter~\cite{1217270} allows for two distinct standard deviations, $\sigma_u$ and $\sigma_v$, along orthogonal axes $(u, v)$ with the kernel rotated by an angle $\phi$. This produces an elongated Gaussian that smooths more along one axis while preserving structures orthogonal to it.
This enhances elongated, line-like structures while suppressing non-linear or noise-like components.

To handle merged lines across adjacent text columns, a vertical distance map is computed by summing the distances from each white pixel to the nearest ink pixels above and below. This map is thresholded to generate a vertical separator mask, which is used to eliminate merging line blobs across columns.
The resulting line blobs are vertically dilated using a structuring element proportional to the estimated character height and intersected with the inverted binary mask to isolate text regions from noise and non-text areas. These refined line blobs are then used as markers to guide a watershed algorithm, which segments the pixels of individual text lines from the detected text regions.

\subsubsection{SRCB: Ding Lei, from Ricoh Software Research Center Beijing Co., Ltd.}

The main idea behind the SRCB method is to recognize text lines using a semantic segmentation model and then apply post-processing techniques to correct common segmentation errors. For segmentation, the authors adopt the SegFormer model~\cite{NEURIPS2021_64f1f27b}, a lightweight and efficient vision transformer that combines a hierarchical Transformer encoder with a simple multi-layer perceptron decoder. This architecture enables high-resolution prediction while remaining computationally efficient, making it suitable for dense pixel-wise tasks like text line segmentation.
In experiments evaluating different loss functions, Dice loss and Lovász-Softmax loss produced comparable results in terms of mean Intersection over Union at the pixel level, but yielded different error characteristics. Specifically, segmentation results trained with Dice loss tended to over-segment text lines, while those trained with Lovász-Softmax loss often under-segmented by merging adjacent lines.
To address the over-connection issue typical of Lovász-Softmax outputs, a simple post-processing method was proposed: (a) Contour detection: Use cv2.findContours to extract the boundaries of connected components; (b) Disconnection points: Identify point pairs that are spatially close (in Euclidean distance) but distant in the contour sequence, indicating a likely erroneous connection between two lines; (c) Separation: Draw a black line at the midpoint of these pairs to separate the lines; (d) Cleanup: Remove overly small components and assign a unique color to each segmented line.
This approach improves the final segmentation results by enforcing clearer line boundaries in densely connected regions.

\begin{figure}[htb]
\centering
\includegraphics[width=1\linewidth]{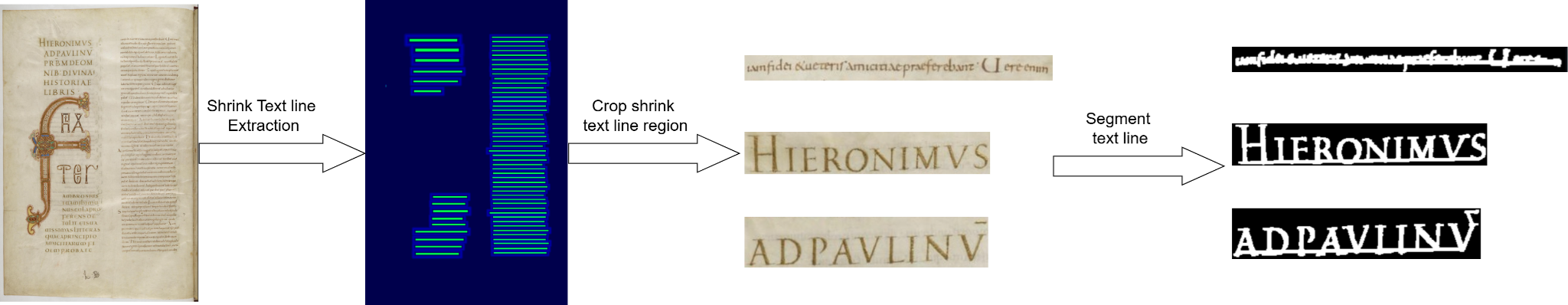} 
    \caption{Text line segmentation pipeline of VAI-OCR team.}%
    \label{vaiocr}%
\end{figure}

\subsubsection{VAI-OCR: Nguyen Nam Quan and Tran Tuan Anh, from Artificial Intelligence department of Viettel AI from Vietnam.}

The proposed approach is designed to extract target masks through a two-stage pipeline (Fig.~\ref{vaiocr}: (a) shrink text line extraction and (b) full text line segmentation. The system builds upon the well-established DBNet framework~\cite{Liao_Wan_Yao_Chen_Bai_2020}, which utilizes shrunk text regions during training and expands them during inference to recover complete line masks. The first stage employs a model that predicts a shrunk mask (with a shrink ratio of 0.4) alongside a threshold mask used to delineate line boundaries. Input images are divided into 224×224 patches. The loss function follows~\cite{Liao_Wan_Yao_Chen_Bai_2020} and combines binary cross-entropy (BCE) and Dice loss for the shrunk mask, as well as L1 loss for the threshold map. The core architecture is a SegFormer model with an EfficientNet-B7 backbone, selected for its balance between accuracy and computational efficiency. During inference, connected components are detected in the shrunk mask and expanded to reconstruct the full text line regions.

In the second stage, each extracted text line is cropped, resized to a fixed height of 64 pixels (maintaining aspect ratio), and padded to match the longest sequence in the batch. The segmentation model is an ensemble of U-Net and SegFormer architectures, with a ResNeXt-50 backbone. The training loss combines BCE and Dice loss. To enhance generalization, the model is pretrained on the DIVA-HisDB dataset~\cite{diva}, using the training and validation folders for training and the public folder for validation. Pretraining is applied to both stages of the pipeline, initialized with ImageNet weights~\cite{5206848}, and followed by few-shot fine-tuning on the target dataset.

\subsubsection{PERO: Martin Kišš, Martin Kostelník, Michal Hradiš, Jan Kohút, Karel Beneš, Marek Vaško, Martin Fajčík, and Martin Dočekal from Brno University of Technology.}
PERO's system is based on a model for document layout analysis, and the entire processing pipeline is divided into two stages: (1) layout analysis and (2) post-processing. In the first stage, a model based on the ParseNet~\cite{PERO} architecture, trained on the PERO layout dataset~\cite{PERO}, is used to detect text lines and text regions. For each document image, pixel-wise predictions are generated for baselines, the distances from each baseline to the top and bottom edges of its corresponding bounding polygon, and the borders of text regions. These outputs are then used to construct individual text lines and group them into text regions.

In the second stage, the localized text lines and text regions are used to produce final segmentation maps. Initially, adaptive Gaussian thresholding is applied within each text region to segment individual lines. To ensure continuity across line segments, the predicted baselines are rendered onto the segmentation maps, thereby bridging fragmented components. Morphological dilation is then applied to enhance the thickness of the segmented lines, and connected components with an area below a predefined threshold are removed. All post-processing operations are performed using implementations from OpenCV.
The provided development data was not used for training; however, it was employed to fine-tune the parameters of the processing pipeline.

\subsubsection{DIA-Group: Hadia Showkat Kawoosa,
Sahana Rangasrinivasan, Srirangaraj Setlur, Venu Govindaraju, and Puneet Goyal from State University of New York at Buffalo, Indian Institute of Technology Ropar and Amrita Vishwa Vidyapeetham.}

The DIA-Group addresses the limitations of training solely on full-page images, which often leads to a lack of fine-grained feature learning, by introducing a patch-based augmentation strategy. For each training sample, C spatially distinct patches are randomly sampled from different regions of the document. These patches are centered on a black canvas and resized to the original resolution (1344×2016), ensuring consistent input dimensions. This method enables the model to learn both global structure and local detail, enhancing generalization across diverse manuscript layouts.

The architecture builds on U-Net++~\cite{unetplus}, incorporating an EfficientNet-B4 encoder to balance performance and computational efficiency. While U-Net offers an encoder-decoder structure with skip connections, U-Net++ enhances feature fusion through nested dense skip pathways, enabling improved multi-scale integration and gradient flow. To offset the increased computational cost of these dense connections, EfficientNet-B4 is used as the encoder. It employs compound scaling and integrates Squeeze-and-Excitation modules to refine channel-wise features via attention mechanisms.

The model outputs are refined by the U-Net++ decoder, which progressively upsamples and reconstructs the segmentation map. Given the presence of noise, irregular layouts, and low contrast in historical documents, this design proves especially effective.
Training is conducted using the Jaccard Loss, which helps address class imbalance common in segmentation tasks. Post-inference, a morphological closing operation is applied to clean up the segmentation output. Using a 7×7 structuring element, dilation followed by erosion smooths irregular boundaries and fills small gaps, improving segmentation quality in challenging document conditions.
Experiments were performed on a high-performance system with four NVIDIA RTX A6000 GPUs and 64GB RAM. The model was trained with the Adam optimizer (learning rate = 0.001, weight decay = 0.00001) for up to 200 epochs. Early stopping was applied if the validation loss did not improve for 50 consecutive epochs after the first 100.

\subsubsection{CV-Lab: Rafael Sterzinger and Florian Kleber from Technische Universität Wien.}

Given the limited availability of training data, the preprocessing pipeline is designed to maximize the
model’s exposure to varied content. During training, each image undergoes light geometric augmentations such as slight rotations and shearing to enhance robustness. From the augmented image, a random 448×448 pixel patch is extracted to serve as input. For validation and testing, the entire image is first padded so that its dimensions align with a uniform patch grid. It is then systematically divided into overlapping 448×448 patches to ensure full coverage. Once predictions are made for each patch, they are systematically recombined to reconstruct the original image. To handle the overlap between patches, a Gaussian weighting function is applied to give more importance to the center of each patch, minimizing artifacts at the edges. The patches are merged by averaging their values based on the accumulated weights from the overlap regions. This process ensures a smooth transition between patches and results in a seamless and accurate reconstruction of the full image, which is then cropped to its original size.

CV-Lab trains a U-Net~\cite{unet} model employing a ResNet-50 encoder pretrained on the ImageNet dataset~\cite{5206848} to leverage strong feature extraction capabilities. To mitigate class imbalance and encourage precise delineation of segmentation boundaries, we use the Dice loss as the sole supervision signal. Training is conducted over 100 epochs with a batch size of 8, using the Adam optimizer.
To further enhance segmentation quality, particularly in terms of structural consistency, the method incorporates the topology-aware loss proposed by Grim et al.~\cite{Grim_Chandrashekar_Sümbül_2025}. Unlike standard voxel-wise losses, this approach penalizes topological errors, such as false merges and splits, by evaluating segmentation quality at a structural level. The hyperparameters are set to $\alpha = 1$ and $\beta = 0.5$, following the original paper, to emphasize penalizing structural inaccuracies.

The training strategy involves first learning a baseline model using the standard Dice loss, followed by fine-tuning the network with a combined loss (Dice + topology-aware) for an additional 100 epochs to refine structural segmentation fidelity.

\subsubsection{BBA: Amine Beghoura, and Mouhoub Belazzoug from University of Bordj Bou Arreridj, Algeria.}
The proposed system presents a structured, two-stage framework based on U-Net architectures~\cite{unet} for the task of text-line segmentation in historical handwritten documents. Traditional segmentation models often struggle in these contexts, primarily due to the lack of structural guidance and the presence of visually ambiguous regions. To address these challenges, the authors introduce a pipeline that first localizes paragraph-level regions and subsequently performs detailed line-level segmentation within these regions. This decomposition into hierarchical subtasks allows the model to narrow its focus progressively, reducing noise and improving accuracy. The first stage uses a U-Net with an encoder-decoder structure to detect paragraph regions. As there are no explicit paragraph annotations in the U-DIADS-TL dataset, synthetic labels are generated by dilating and eroding the ground-truth line masks using morphological operations. These surrogate masks group nearby lines into coherent blocks while filtering out fine-scale artifacts. Prior to feeding the images into the U-Net, Sauvola binarization~\cite{SAUVOLA} is applied to counteract illumination and ink variability, producing cleaner input representations. The encoder captures coarse layout features while the decoder reconstructs the spatial map of paragraphs, aided by skip connections that preserve resolution and edge details. The result is a coarse but reliable paragraph segmentation map that isolates textual content from background clutter.

In the second stage, the output paragraph mask is leveraged as a spatial prior to guide a second U-Net tasked with high-resolution text-line segmentation. This is accomplished through an attention-gating mechanism that enhances the input to the second U-Net by emphasizing relevant regions. Specifically, a lightweight convolutional subnetwork processes the paragraph mask to produce a soft attention map that assigns each pixel a probability of belonging to a text-bearing region. This map is then applied multiplicatively to the original grayscale input, suppressing noise and irrelevant details while preserving the fidelity of textual regions. This preprocessing step ensures that the second U-Net concentrates computational resources on areas likely to contain line structures. Architecturally similar to the first network, the second U-Net features an encoder that extracts fine-grained features—such as stroke continuity, inter-line spacing, and line boundaries—and a decoder that reconstructs a pixel-wise segmentation mask of individual text lines. The inclusion of attention-guided input enhances the precision of segmentation boundaries, especially in areas with dense or overlapping text.

\subsubsection{GPI: Ignacio Ramirez from Grupo de Procesamiento de Imágenes, Universidad de la República, Uruguay.}
The approach presented by the GPI group constitutes a deliberately non-neural, traditional image processing pipeline that serves as a proof-of-concept for the efficacy of classical techniques in document layout analysis. The approach is fully deterministic, with the sole learned parameter being the maximum area of a connected component, which is empirically derived from the training data. The method operates in three stages: pre-processing, detection, and post-processing. 

During pre-processing, input document images are converted to grayscale and subjected to denoising via Total Variation minimization, followed by contrast enhancement using a morphological Top-Hat transform with a circular structuring element. These last two steps are performed using Total Variation Minimization~\cite{Totalvariation}. This step yields smoothed yet structurally coherent text images and suppresses background noise, improving the reliability of subsequent segmentation stages.

In the detection phase, a custom blurring operation is applied to the pre-processed image using an elongated kernel that suppresses ascenders and descenders while preserving the main horizontal text structures. This enables robust extraction of vertical intensity profiles to detect columnar layouts, which are used to divide the image into vertical bands for independent processing. Line detection then applies Otsu’s thresholding~\cite{otsu} to the blurred image, producing binary blobs corresponding to text components. For each blob, a second-order polynomial is fitted to the horizontal midpoints, capturing the curvature of the text lines. These curves are extrapolated to bridge fragmented regions and subsequently dilated with a large, anisotropic ellipsoidal kernel to form an initial segmentation mask.

Post-processing refines this mask to better align with the labeling criteria of the ground truth, which includes the full spatial extent of character shapes. A secondary thresholding step is applied to the contrast-normalized image using Otsu’s method, and each resulting connected component is matched to the closest region in the initial segmentation via maximal intersection. Components without intersection or exceeding an empirically determined size threshold set at 120\% of the 95th percentile of training-set component areas are discarded to reduce over-segmentation and false positives. This results in a final segmentation output that is both semantically accurate and visually aligned with human-labeled ground truth, demonstrating that competitive performance in line segmentation can be achieved without deep learning by carefully engineering classical vision algorithms.

\subsubsection{LDLD: AmirHosein Komeili, Boshra Rajaei, and Jean-Michel Morel from BSadjad University of Technology and ENS Paris-Saclay.}
The authors developed a multi-stage algorithm, referred to as LDLD (Layout-based Document Line Detector), which begins with document layout analysis to separate various components such as text blocks, images, tables, and others.
The LDLD algorithm comprises the following stages:
\begin{itemize}
    \item Layout Segmentation: RTMDet (Real-Time Models for Object Detection) is fine-tuned on the PrimaLayout dataset~\cite{5277696} and the FEST training and validation datasets. RTMDet~\cite{lyu2022rtmdetempiricalstudydesigning} is an object detection model designed to balance accuracy and efficiency. It features a unified architecture with explicit vision-centric pretraining, enabling the model to better capture spatial structures within images.
    \item Layout Post-processing: This includes: (a) Grouping bounding boxes based on Intersection over Union (IoU); (b) Merging bounding boxes within each group; (c) Removing blank areas within bounding boxes; (d) Eliminating bounding boxes that overlap excessively with larger ones based on a predefined threshold.
    \item Document Line Segmentation: YOLOv11~\cite{yolo} is trained on 142 training and 35 validation images randomly sampled from multiple public datasets. Due to the lack of appropriate ground truth for the FEST train and validation sets, these were not included in this stage.
    \item Line Segmentation Post-processing: Following mask prediction for each document line, overlapping issues are addressed by removing small regions with high overlap with larger ones, as well as eliminating small overlaps between neighboring masks.
    \item Line Detection via Regression: The final line positions are estimated using regression. For vertically aligned lines, the regression variables are adjusted accordingly.
\end{itemize}

\begin{table}[htb]
\huge
\centering
\def\arraystretch{1.2}
\setlength{\tabcolsep}{0.25em}
\resizebox{\textwidth}{!}{
\begin{tabular}{l|ccccc|ccccc|ccccc}
\toprule
\textbf{Backbone} & 
\multicolumn{5}{c|}{\textbf{Latin 2}} & 
\multicolumn{5}{c|}{\textbf{Latin 14396}} & 
\multicolumn{5}{c}{\textbf{Syriaque 341}} \\
\midrule
& PIU & LIU & DR & RA & FM 
& PIU & LIU & DR & RA & FM 
& PIU & LIU & DR & RA & FM \\
\midrule

FCN & 0.51 & 0.45 & 0.29 & 0.17 & 0.21
    & 0.55 & 0.49 & 0.44 & 0.27 & 0.33
    & 0.18 & 0.08 & 0.03 & 0.04 & 0.03 \\
    
PSPNet & 0.52 & 0.40 & 0.15 & 0.11 & 0.13
       & 0.50 & 0.52 & 0.34 & 0.30 & 0.32
       & 0.08 & 0.02 & 0 & 0 & 0\\
      
DeepLabV3+ & 0.55 & 0.53 & 0.28 & 0.15 & 0.19
          & 0.57 & 0.58 & 0.51 & 0.33 & 0.40
          & 0.34 & 0.22 & 0.09 & 0.05 & 0.06\\
\midrule           
CV-Group & \textbf{0.83} & 0.96 & \textbf{0.87} & 0.18 & 0.30
       & 0.67 & 0.77 & 0.63 & 0.14 & 0.22
       & 0.85 & \textbf{0.97} & \textbf{0.96} & 0.19 & 0.31\\

Codecrackers & 0.52	&0.13	&0.03	&0.02	&0.02&	0.53&	0.47&	0.23&	0.18&	0.20&
0.26&	0.01&	0&	0&	0\\

TAU-CH & 0.71 & 0.86 & 0.68 & 0.74 & 0.70
       & 0.65 & 0.61 & 0.19 & 0.17 & 0.18
       & 0.72 & 0.90 & 0.63 & 0.69 & 0.66 \\

SRCB & 0.78 & 0.94 & 0.85 & 0.78 & 0.81
       & 0.75 & 0.90 & \textbf{0.75} & \textbf{0.73} & \textbf{0.74}
       & 0.81 & 0.94 & 0.92 & 0.88 & 0.90 \\

VAI-OCR & 0.72 & 0.83 & 0.79 & 0.84 & 0.81
       & 0.55 & 0.65 & 0.58 & 0.70 & 0.64
       & 0.67 & 0.76 & 0.75 & 0.84 & 0.79\\

PERO & 0.79 & \textbf{0.97} & 0.80 & \textbf{0.87} & \textbf{0.83}
       & \textbf{0.76} & \textbf{0.94} & 0.70 & 0.72 & 0.71
       & \textbf{0.86} & \textbf{0.97} & 0.94 & \textbf{0.96} & \textbf{0.95} \\

DIA-Group & 0.65 & 0.72 & 0.58 & 0.39 & 0.46
       & 0.31 & 0.22 & 0.15 & 0.17 & 0.16
       & 0.67 & 0.78 & 0.74 & 0.62 & 0.67 \\

CV-Lab & 0.73 & 0.80 & 0.78 & 0.44 & 0.56
       & 0.58 & 0.79 & 0.59 & 0.42 & 0.49
       & 0.74 & 0.79 & 0.79 & 0.55 & 0.64\\

BBA & 0.52 & 0.26 & 0.23 & 0.12 & 0.16
       & 0.47 & 0.09 & 0.05 & 0.03 & 0.04
       & 0.56 & 0.31 & 0.30 & 0.15 & 0.20 \\

GPI & 0.78 & 0.93 & 0.73 & 0.79 & 0.75
       & 0.71 & 0.76 & 0.34 & 0.34 & 0.34
       & 0.74 & 0.88 & 0.62 & 0.59 & 0.60 \\

LDLD & 0.35 & 0. & 0. & 0. & 0.
       & 0.39 & 0. & 0. & 0. & 0.
       & 0.42 & 0. & 0. & 0. & 0. \\

\bottomrule
\end{tabular}
}
\caption{Performance comparison of the different systems on the U-DIADS-TL dataset. For each manuscript, results are reported across five metrics (Pixel IU, Line IU, DR, RA, FM). The best performances are highlighted in bold.}
\label{tableFS}
\end{table}

\section{Evaluation Protocol}
The evaluation phase assessed participant systems on 15 private, unpublished images per manuscript, for which only the original images, without the corresponding GTs, were provided to the competitors. Evaluation followed established practices by relying on the two most prominent evaluation benchmarks available in the literature~\cite{diva,eval}, which delineated five key metrics: Pixel Intersection over Union (PIU), Line Intersection over Union (LIU), Detection Rate, Recognition Accuracy, and F-measure.

PIU and LIU are based on the standard Intersection over Union (IU) metric, defined as:
\begin{equation}
\text{IU} = \frac{\text{TP}}{\text{TP} + \text{FP} + \text{FN}}
\end{equation}
where TP, FP and FN denote True Positives, False Positives, and  False Negatives, respectively.

PIU evaluates the segmentation accuracy at the pixel level. Here, TP corresponds to correctly predicted text line pixels, FP to background pixels incorrectly labeled as text, and FN to missed text pixels.

LIU evaluates segmentation at the line instance level. 
TP represents correctly detected lines, FP represents extra (false) lines, and FN represents missed lines.
To establish matches between predicted and ground truth lines, a component-wise overlap is computed. A match is confirmed if both pixel-level precision and recall exceed a threshold of 75\%. 

Detection Rate (DR), Recognition Accuracy (RA) and F-Measure(FM) are based on the MatchScore. For a given image, let $R_i$ be the set of points inside the $i$th detected line segment, $G_j$ be the set of points inside the $j$th line segment in the ground truth, and $T(p)$ be a function that counts the points in a given set $p$, then $\text{MatchScore}(i, j)$ is calculated as follows:
\begin{equation}
\text{MatchScore}(i, j) = \frac{T(G_j \cap R_i)}{T(G_j \cup R_i)}
\end{equation}
A region pair $(i, j)$ is considered a one-to-one match if the $\text{MatchScore}(i, j)$ is equal to or above the threshold $T$. In this competition, $T = 75\%$, which in our experience is a value that allows to achieve good quality segmentations, without being too strict.
Let $N_1$ and $N_2$ be the number of ground-truth and detected lines, respectively, and let $M$ be the number of one-to-one matches. The tool calculates DR, RA, and FM as follows:

\begin{equation}
\text{DR} = \frac{M}{N_1}, \quad \text{RA} = \frac{M}{N_2}, \quad \text{FM} = \frac{2 \cdot \text{DR} \cdot \text{RA}}{\text{DR} + \text{RA}}
\end{equation}

All metrics are computed individually for each manuscript. The final ranking on the leaderboard is based on the average Line IU score across the three manuscripts provided in the evaluation phase.

\begin{table}[htb]
\small
\centering
\def\arraystretch{1.2}
\setlength{\tabcolsep}{0.35em}
\resizebox{.8\textwidth}{!}{
\begin{tabular}{lcccccc}
\cline{1-7}
                                                & \multicolumn{5}{c}{Average}                                                        \\ \hline
                                          &Rank      & PIU      & LIU       & DR      & RA      & FM      \\ \hline
PERO                                      &1      & \textbf{0.803} & \textbf{0.961} & 0.815          & \textbf{0.849} & \textbf{0.830} \\
SRCB                                       &2     & 0.782          & 0.924          & \textbf{0.839} & 0.797          & 0.817          \\
CV-Group                                   &3     & 0.784          & 0.897          & \textbf{0.818} & 0.167          & 0.277          \\
GPI              &4   & 0.743          & 0.856          & 0.561          & 0.575          & 0.566          \\
CV-Lab             &5     & 0.682          & 0.792          & 0.720          & 0.470          & 0.601          \\
TAU-CH          &6        & 0.691          & 0.790          & 0.501          & 0.533          & 0.514          \\
VAI-OCR       &7  & 0.649          & 0.747          & 0.709          & 0.795          & 0.748          \\
DIA-Group                    &8                   & 0.541          & 0.572          & 0.488          & 0.392          & 0.430          \\

BBA &9& 0.516          & 0.219          & 0.195          & 0.098          & 0.130          \\
Codecrackers          &10  & 0.434	&0.203&	0.087	&0.064&	0.072          \\
LDLD       &11     & 0.384          & 0.000          & 0.000          & 0.000          & 0.000          \\

 \hline
\end{tabular}}
\caption{Final ranking of the competition, defined by the score obtained by the systems on the Line IU metric. The best-performing method for each metric is highlighted in bold.}\label{tab:ranking}
\end{table}§

\begin{figure}[htb]
\centering
\includegraphics[width=.8\linewidth]{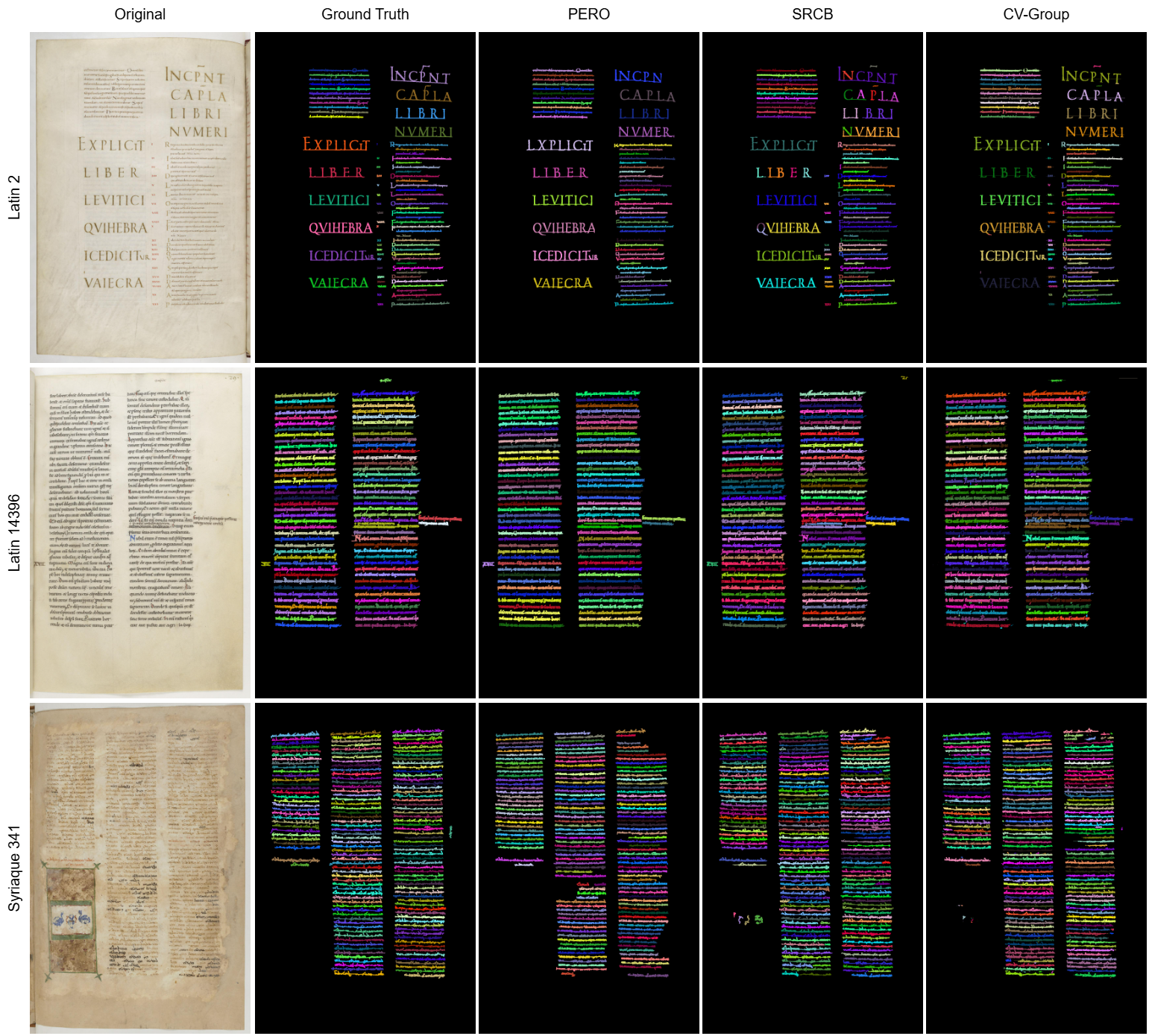} 
    \caption{Qualitative comparison of the top three performing systems for text line segmentation.}%
    \label{risultati}%
\end{figure}

\section{Experimental Results}
Table~\ref{tableFS} presents the performance of the different models submitted for the FEST competition on the U-DIADS-TL dataset. Alongside the results of the competing systems, we also report the performance of three popular deep-learning-based semantic segmentation models (DeepLabV3+~\cite{deeplabv3plus}, FCN~\cite{fcn}, and PSPNet~\cite{pspnet}), as baseline references, following the benchmarks proposed in~\cite{ircdl}. 
The results are reported separately for each manuscript, with the best scores highlighted in bold. As observed, most of the competition entries outperform the baseline models.

Table~\ref{tab:ranking} reports the final ranking of the competition, based on the Line IoU metric used to evaluate system performance. Notably, the best-performing system, submitted by the PERO team, achieved the top scores across all evaluation metrics, except for the Detection Rate metric. This system reached an average Line IoU score of 0.961 across the dataset, outperforming the third-best performing system, submitted by the CV-Group team, by 6.4\%, and the second-place system from the SRCB team by 3.7\%.

Notably many of the partecipants' systems, including the one proposed by the third placed CV-Group, achieved lower performance on the RA and FM metrics, suggesting that there is still space to explore more advanced approaches in this field. 
Furthermore, a special mention should go to the GPI group, which managed to achieve surprisingly good performance, as confirmed by its fourth place in the ranking, by relying exclusively on traditional computer vision techniques, without the use of any neural-based component.
Finally, Figure~\ref{risultati} presents the qualitative results of the top three participating systems. For each manuscript class in the dataset, we showcase a representative segmentation map generated by each system and compare it against the corresponding ground truth mask.

\section{Conclusion}
In this paper, we introduced the Competition on FEw-Shot Text line segmentation of ancient handwritten documents (FEST), part of ICDAR 2025, which tackles text line segmentation under a few-shot learning setting. Unlike traditional methods requiring extensive annotations, FEST challenges participants to design models that generalize across diverse historical scripts using only a few labeled samples per manuscript.
By framing text line segmentation as a few-shot problem, FEST bridges the gap between real-world constraints and state-of-the-art machine learning techniques, encouraging innovation in areas such as domain adaptation and low-resource document analysis.

\section*{Declarations}
Partial financial support was received from: PNRR DD 3277 del 30 dicembre 2021 (PNRR Missione 4, Componente 2, Investimento 1.5) - iNEST; Strategic Departmental Plan on Artificial Intelligence, University of Udine; Strategic Departmental Plan and interdepartmental center AI4CH – Artificial Intelligence for Cultural Heritage; dall’Unione europea - Next Generation EU, Missione 4 Componente 2 CUP G23C24000790006 – Progetto COVERT.

\bibliographystyle{splncs04}
\bibliography{mybibliography}

\end{document}